\pdfoutput=1

\documentclass[11pt]{article}

\usepackage[final]{coling}

\usepackage{times}
\usepackage{latexsym}
\usepackage[T1]{fontenc}
\usepackage{mathptmx} 
\usepackage[utf8]{inputenc}
\usepackage{microtype}
\usepackage{inconsolata}
\usepackage{graphicx}
\usepackage{algorithm}
\usepackage{algpseudocode}
\usepackage{booktabs}
\usepackage{multirow}
\usepackage{multicol}
\usepackage{marvosym}
\usepackage{printlen}
\usepackage{hyperref}
\usepackage{url}
\usepackage[standard]{ntheorem}
\usepackage{wrapfig}
\usepackage{caption}
\usepackage{arydshln}
\usepackage{colortbl} 
\usepackage{amsmath}
\usepackage{times}
\usepackage{latexsym}
\usepackage{listings}

\title{Who Wrote This? The Key to Zero-Shot LLM-Generated \\Text Detection Is \textsc{GECScore}}

\newcommand\blfootnote[1]{%
  \begingroup
  \renewcommand\thefootnote{}\footnote{#1}%
  \addtocounter{footnote}{-1}%
  \endgroup
}

\author{Junchao Wu$^{\clubsuit}$~~~
        Runzhe Zhan$^{\clubsuit}$~~~
        Derek F. Wong$^{\clubsuit\text{~\Letter}}$~~~
        Shu Yang$^{\clubsuit}$~~~  \\
        \textbf{Xuebo Liu$^{\spadesuit}$}~~~  
        \textbf{Lidia S. Chao}$^{\clubsuit}$~~~
        \textbf{Min Zhang}$^{\spadesuit}$~~~ \\
    $^{\clubsuit}$NLP$^2$CT Lab, Department of Computer and Information Science, University of Macau \\
      \texttt{nlp2ct.\{junchao,runzhe, shuyang\}@gmail.com, \{derekfw,lidiasc\}@um.edu.mo} \\
    $^{\spadesuit}$Institute of Computing and Intelligence, Harbin Institute of Technology, Shenzhen, China \\
        \texttt{\{liuxuebo, zhangmin2021\}@hit.edu.cn}
    }

\begin{document}
\maketitle
\blfootnote{$^\text{\Letter}$~Corresponding author.}

\begin{abstract}
The efficacy of detectors for texts generated by large language models (LLMs) substantially depends on the availability of large-scale training data. However, white-box zero-shot detectors, which require no such data, are limited by the accessibility of the source model of the LLM-generated text. In this paper, we propose a simple yet effective black-box zero-shot detection approach based on the observation that, from the perspective of LLMs, human-written texts typically contain more grammatical errors than LLM-generated texts. This approach involves calculating the \textbf{G}rammar \textbf{E}rror \textbf{C}orrection \textbf{Score} (\textsc{GECScore}) for the given text to differentiate between human-written and LLM-generated text. Experimental results show that our method outperforms current state-of-the-art (SOTA) zero-shot and supervised methods, achieving an average AUROC of 98.62\% across XSum and Writing Prompts dataset. Additionally, our approach demonstrates strong reliability in the wild, exhibiting robust generalization and resistance to paraphrasing attacks. Data and code are available at: \url{https://github.com/NLP2CT/GECScore}.

\end{abstract}

\section{Introduction}

The history of zero-shot methods for detecting LLM-generated text can be traced back to research on machine translation (MT) detection. These studies utilized linguistic features to determine whether a given text was generated by MT systems \cite{DBLP:conf/acl/Corston-OliverGB01,DBLP:conf/acl/AraseZ13}. Progress in the field has led to the development of a suite of zero-shot methods specifically designed for LLM-generated text detection. These methods base their calculations on the logits that are generated by the source model, including approaches such as Log-Likelihood \cite{DBLP:journals/corr/abs-1908-09203}, Perplexity \cite{DBLP:conf/nldb/Beresneva16, DBLP:journals/corr/abs-2305-18226}, Entropy \cite{DBLP:conf/ecai/LavergneUY08}, Log-Rank \cite{DBLP:conf/acl/GehrmannSR19}, and the Log-Likelihood Log-Rank Ratio (LRR) \cite{DBLP:conf/emnlp/SuZ0N23}. Perturbed-based zero-shot detection methods, including DetectGPT \cite{DBLP:conf/icml/Mitchell0KMF23} and Normalized Log-Rank Perturbation (NPR) \cite{DBLP:conf/emnlp/SuZ0N23}, calculate the Log-Likelihood and Log-Rank curvature on the perturbed texts to distinguish between the two types of texts. More recently, \citet{DBLP:journals/corr/abs-2310-05130} replaced the perturbation step of DetectGPT with a more efficient sampling step, achieving better performance and speed.

\begin{figure}[t]
    \centering
    \includegraphics[width=0.48\textwidth]{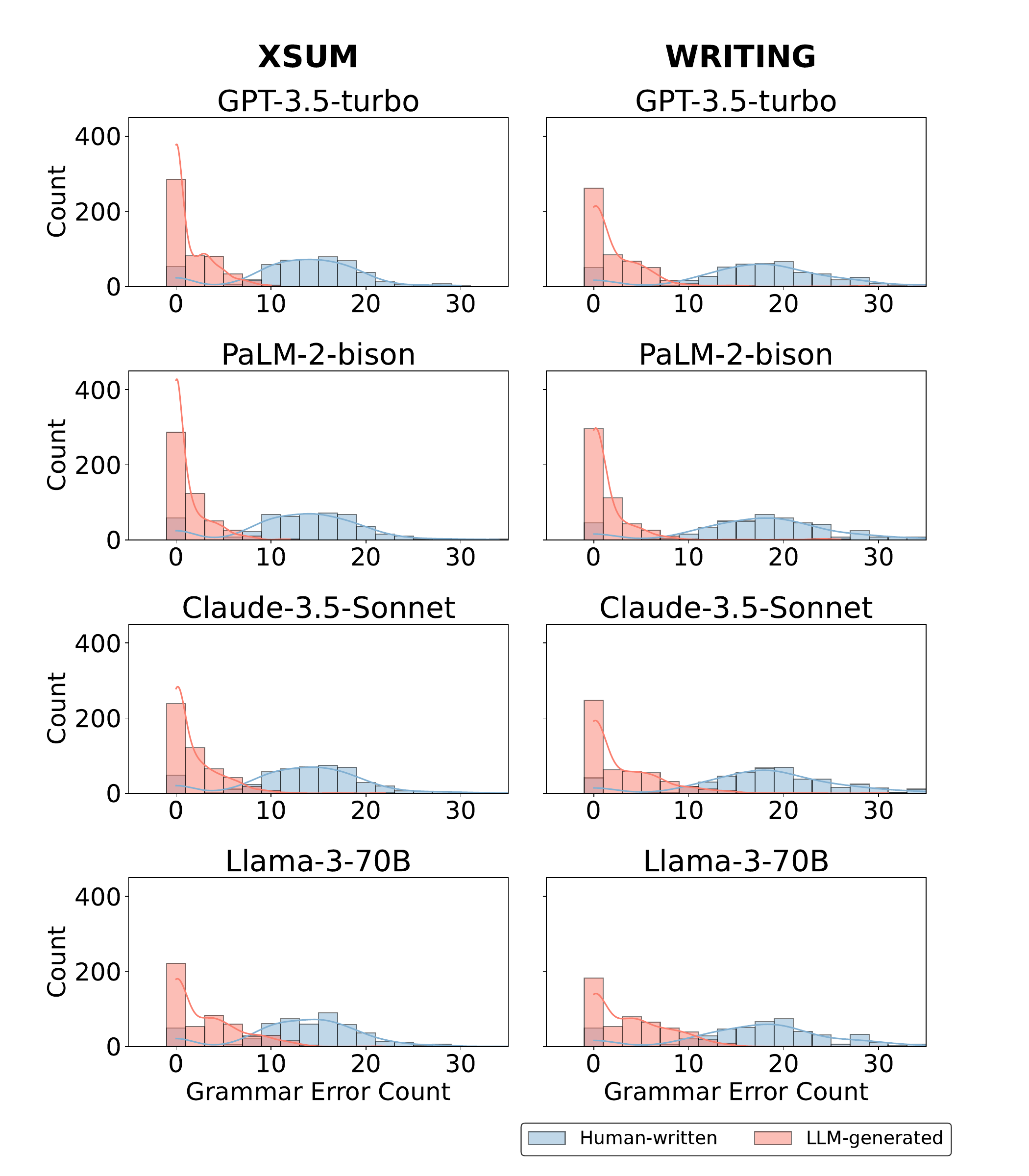}
    \caption{
    Distribution of \textit{grammar errors} of human-written texts and LLM-generated texts by GPT-3.5-turbo, PaLM-2-bison, Claude-3.5-Sonnet, and Llama-3-70B on XSum and Writing Prompts. GPT-4o~\cite{blog2024gpt4o} is employed for grammar errors marking.}
    \label{fig:Grammar_Error_Count_Difference}
\end{figure}

However, current zero-shot detectors are not reliable enough when dealing with cleverly revised text, such as paraphrased text \cite{DBLP:journals/corr/abs-2303-11156,DBLP:journals/corr/abs-2303-13408,DBLP:journals/corr/abs-2410-23746}. They also struggle to detect text from unknown LLMs \cite{DBLP:journals/corr/abs-2310-05130}. While supervised approaches based on fine-tuning pre-trained models (PLMs) demonstrate enhanced performance, they often fail in out-of-distribution settings \cite{DBLP:journals/corr/abs-2305-13242} and are susceptible to attack through targeted training of an LLM \cite{nicks2023language}. Given these challenges, we focus on a more challenging but practical direction. We explore a black-box zero-shot approach that does not rely on access to the source model, extensive training datasets, or fine-tuning PLMs as classifiers.

In this study, we propose a simple but effective zero-shot detection method that leverages an external grammar error correction model. By computing the \textsc{GECScore} for the given text, our approach distinguishes between human-written and LLM-generated text. Extensive experimental results show that our method not only outperforms current SOTA zero-shot and supervised techniques, but also shows remarkable reliability in the wild, including robust generalization capability and resistance to paraphrasing attack.
\section{Related Works}
\subsection{White-box LLM-generated Text Detection}

The white-box LLM-generated text detection methods require access to the source model. Current white-box approaches typically employ zero-shot techniques, where specific metrics are derived from the logits of the LLM output. These metrics are then compared against a statistically derived threshold, which serves as the criterion for identifying LLM-generated text. The most commonly used metric is Log-Likelihood \cite{DBLP:journals/corr/abs-1908-09203}, which evaluates whether a given text was generated by an LLM by measuring the average token-wise log probability of each token. Similarly, Rank \cite{DBLP:conf/acl/GehrmannSR19} calculates the average of the absolute rank and entropy values for each token. Log-Rank \cite{DBLP:conf/acl/GehrmannSR19} further improves on Rank by applying a logarithmic transformation to the rank values of each token, achieving better performance. A notable method is LRR~ \cite{DBLP:conf/emnlp/SuZ0N23}, which combines Log-Likelihood and Log-Rank. By taking the ratio of these metrics, LRR comprehensively incorporates their strengths for improved performance.

Certain white-box methods perform LLM-generated text detection by quantifying metric variations following perturbations. DetectGPT \cite{DBLP:conf/icml/Mitchell0KMF23} distinguishes LLM-generated text by introducing perturbations to the original text via a T5 model \cite{DBLP:journals/jmlr/RaffelSRLNMZLL20} and measuring the resultant alterations in the log probability of the source LLM. This method capitalizes on the observation that the average decrease in Log-Likelihood of perturbed LLM-generated text consistently exceeds that of perturbed human-written text. Another similar and concurrent work by \citet{DBLP:conf/emnlp/SuZ0N23} performs the detection task by measuring Log-Rank. Although perturbed-based curve statistical methods exhibit superior capabilities compared to conventional zero-shot methods, the time overhead of perturbation greatly reduces the usability of these methods. In response to this limitation, Fast-DetectGPT \cite{DBLP:journals/corr/abs-2310-05130} replaced the perturbation step of DetectGPT with a more efficient sampling step to achieve superior efficacy and efficiency in LLM-generated-text detection performance.

\begin{figure*}[!ht]
    \centering
    \includegraphics[width=\textwidth, trim=0 0 0 0]{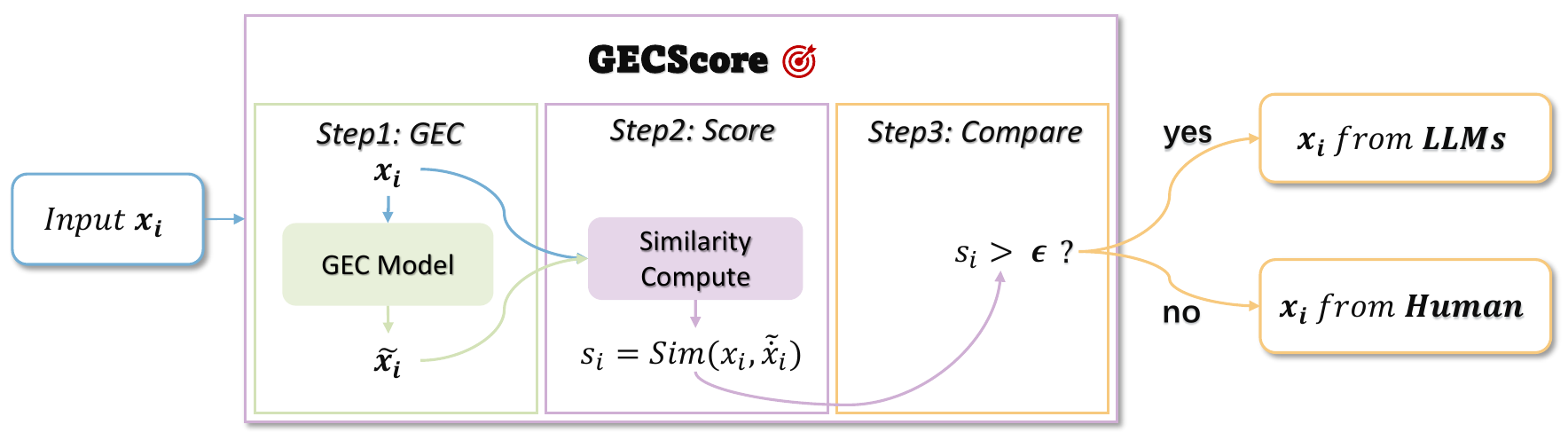}
    \caption{\textsc{GECScore} Framework Overview. First, a grammar correction model generates a grammatically corrected version $\tilde{x_i}$ of the input text $x_i$. Next, the similarity score $s_{i}$ between $\tilde{x_i}$ and $x_i$ is calculated using a similarity metric $Sim$. Finally, if $s_{i}$ meets or exceeds the threshold $\epsilon$, the text is more likely to be generated by LLM.}
    \label{fig:GECScore}
\end{figure*}

\subsection{Black-box LLM-generated Text Detection}

Black-box LLM-generated text detection mainly identifies LLM-generated text by training neural-based classifiers. Fine-tuning PLM-based classifiers is the most widely used black-box method and typically requires a large number of human-written and LLM-generated training samples. Pre-trained language models, such as BERT \cite{DBLP:conf/naacl/DevlinCLT19}, RoBERTa \cite{DBLP:journals/corr/abs-1907-11692}, and XLNet \cite{DBLP:conf/nips/YangDYCSL19}, demonstrate strong semantic understanding ability and effectively enhance the performance of NLP tasks \cite{DBLP:journals/corr/abs-2003-08271}, especially text classification. Previous research has demonstrated that the fine-tuned classifiers based on RoBERTa are highly effective at detecting LLM-generated text \cite{DBLP:journals/corr/abs-2301-07597,DBLP:journals/corr/abs-2304-07666,DBLP:journals/corr/abs-2306-05524,DBLP:journals/corr/abs-2305-07969}. Notably, these fine-tuned classifiers have been shown to outperform other methodologies, such as zero-shot approaches and watermark technologies, while also demonstrating commendable adversarial robustness.

Recently, \citet{DBLP:conf/emnlp/ZhuYCCFHDL0023} demonstrated a novel black-box zero-shot method for detecting LLM-generated text. This method employs BARTScore \cite{DBLP:conf/nips/YuanNL21}, a metric for calculating the semantic similarity between ChatGPT-revised text and the original text. The foundational hypothesis is that ChatGPT tends to introduce fewer modifications to LLM-generated text compared to text written by humans.
\section{Methodology: \textsc{GECScore}}

\subsection{Hypothesis and Motivation} \label{sec:hypo}

The proposed \textsc{GECScore} is based on the hypothesis that, from the perspective of LLMs, human-written texts typically contain more grammatical errors than LLM-generated texts. In other words, when an LLM is used to perform grammatical error correction on an initial text $x_i$, the similarity between the corrected text $\tilde{x_i}$ and the original text $x_i$ should be higher if the text was generated by an LLM, compared to if it was written by a human. This hypothesis relies on two specific features: \textbf{Differences in Grammatical Errors} and \textbf{Differences in LLM Correction Preferences}.

\paragraph{Human-written texts inherently contain more grammatical errors.} The intrinsic complexity of the human brain, combined with psychological and environmental variables, naturally results in a higher error rate in human-written texts compared to LLM-generated outputs. The \textit{Working Memory Theory}~\cite{baddeley1992working,olive2004working} provides an initial explanation for this phenomenon: writing is a complex, high-level cognitive task that relies on working memory to organize ideas, construct sentences, and retrieve linguistic rules. The limitations of working memory often lead to writing errors. 

Research in neuroscience and cognitive psychology, particularly on \textit{Word Priming}~\cite{meyer1971facilitation,neely1977semantic} and \textit{Memory Formation}~\cite{hebb2005organization}, has uncovered the cognitive mechanisms behind linguistic errors. Such mistakes often stem from the brain's tendency to prioritize semantic coherence and narrative fluency over character-level and syntactic accuracy. This tendency is evident even among skilled writers and contributes to the difference in error frequency between human-written and LLM-generated texts. Moreover, theories like \textit{Language Interference} \cite{odlin2003cross}, which points out that multilingual individuals might confuse different linguistic systems and increase the chance of errors, \textit{Attention Bias} \cite{fernandes2019effect,limpo2017examining}, which suggests that distractions can split a writer's focus and raise the possibility of mistakes, and \textit{Cognitive Load} \cite{sweller1988cognitive,kellogg1987writing}, which emphasizes the intense mental effort required for writing tasks (e.g. organizing information) can lead to more errors, all offer further empirical support for the observed differential in error propensity between human-written and LLM-generated texts.

\paragraph{LLMs tend to perform grammatical error correction on human-written texts.} LLMs are generally more familiar with the statistical patterns of texts generated by themselves or other LLMs~\cite{DBLP:conf/emnlp/ZhuYCCFHDL0023}. As a result, they are less sensitive to grammatical errors in such texts and less inclined to correct them, leading to fewer detected and corrected errors. In contrast, LLMs exhibit a stronger tendency to correct human-written texts due to the differences in language style and statistical patterns between human-written texts and the rules learned by LLMs during training. Specifically, informal grammar, stylistic variations, or specific linguistic habits in human-written texts are more likely to be judged as ``grammatical errors'' by LLMs. When correcting human-written texts, LLMs typically adhere more strictly to their internal language modeling norms to identify and address grammatical issues. Consequently, LLMs' correction preferences may amplify the grammatical error differences between LLM-generated texts and human-written texts in grammar correction tasks.

\subsection{Hypothesis Validation}
\label{sec:hypothesis_validation}

\begin{figure}[!ht]
    \centering
    \includegraphics[width=0.42\textwidth]{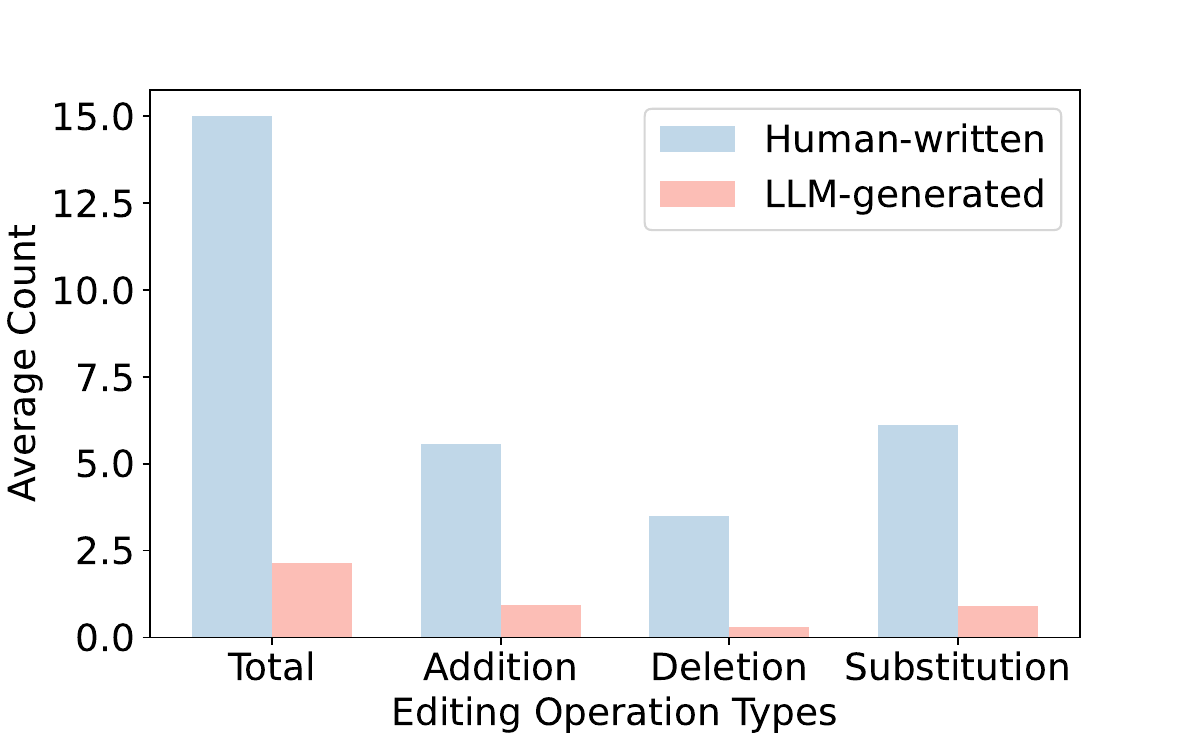}
    \caption{Distribution of \textit{types of editing operations} required for grammar error corrections on human-written texts and LLM-generated texts by GPT-3.5-turbo, PaLM-2-bison, Claude-3.5-Sonnet, and Llama-3-70B on XSum and Writing Prompts.}
    \label{fig:Grammar_Error_Type_comparison}
\end{figure}

We validate the proposed hypothesis. Specifically, we use GPT-4o~\cite{blog2024gpt4o} to annotate grammatical errors in human-written texts and LLM-generated texts produced by GPT-3.5-turbo~\cite{blog2022chatgpt}, PaLM-2-bison~\cite{blog2023palm}, Claude-3.5-Sonnet~\cite{blog2024claude3.5}, and Llama-3-70B~\cite{blog2024llama3} on the XSum and Writing Prompts datasets. We then compile statistics on the total number of grammatical errors in each text, as well as the number of editing operations of addition, deletion, and substitution required for grammatical error correction. Detailed experimental settings and parameters are provided in Appendix~\ref{sec:hypothesis_validation_settings}.
Our findings, illustrated in \autoref{fig:Grammar_Error_Count_Difference}, support the hypothesis across diverse LLMs. Specifically, the number of grammar correction edits per human-written text typically ranges from 10 to 30, while for LLM-generated samples, it ranges from 0 to 10 and is often close to 0. This phenomenon is consistent across both formal and informal writing, although human-written texts in formal writing (XSum) contain relatively few grammatical errors. Furthermore, \autoref{fig:Grammar_Error_Type_comparison} provides a detailed comparison of the types of editing operations required for both text types. We analyze and attribute these differences as follows:

\paragraph{Higher initial quality of LLM-generated texts.} LLM-generated texts typically require fewer additions and deletions, indicating higher initial quality. This is due to extensive training on diverse datasets, which helps LLMs produce grammatically correct and well-structured sentences from the start.

\paragraph{Human-written texts and revision needs.} Human-written texts often involve more additions and deletions. This indicates that human-written drafts may contain more redundant information or omissions, necessitating revisions for quality improvement. Since humans are more prone to oversights or mistakes, especially during rapid drafting.

\paragraph{Consistency in vocabulary choice of LLMs.} Since LLMs are trained on large-scale data and operate probabilistically, they consistently select vocabulary appropriate for the context, typically requiring fewer word substitutions. In contrast, human-written texts, influenced by personal habits and variations in language usage, often require more substitutions to optimize vocabulary usages.

\subsection{Detection Using \textsc{GECScore}}

\begin{algorithm}
\caption{\textsc{GECScore}-based Detection}
\label{GECScore_based_Detection}
\begin{algorithmic}[1]
\Require Sample $x_i$, preliminary sample set $X$, grammar error correction model $g$, similarity function $\mathrm{Sim}$.
\State Expand sample set $X \leftarrow X \cup \{x_i\}$. \Comment{Step 1}
\State Prepare corrected sample set $\tilde{X} \leftarrow \emptyset$. \Comment{Step 2} \\
\textbf{for} each $x_j$ in $X$ \textbf{do}
\State \hspace{2em} $\tilde{x_j} \leftarrow g(x_j)$ 
\State \hspace{2em} $\tilde{X} \leftarrow \tilde{X} \cup \{\tilde{x_j}\}$ 
\State \textbf{end}
\State Initialize similarity scores set ${S} \leftarrow \emptyset$. \Comment{Step 3}
\textbf{for} each pair $(x_j, \tilde{x_j})$ in $\langle X, \tilde{X} \rangle$ \textbf{do}
\State \hspace{2em} ${s}_{j} \leftarrow \mathrm{Sim}(x_j, \tilde{x_j})$
\State \hspace{2em} ${S} \leftarrow {S} \cup \{{s_j}\}$
\State \textbf{end}
\State Calculate and adjust threshold $\epsilon$
\State $\epsilon = \arg \max_{\epsilon'} \left( \text{TPR}(\epsilon') + (1 - \text{FPR}(\epsilon')) \right)$ 
\State Compare with threshold $\epsilon$. \Comment{Step 5}
\State \textbf{if} $s_{i} > \epsilon$ \textbf{then}
\State \hspace{2em} \textbf{return} {True}
\State \textbf{else}
\State \hspace{2em} \textbf{return} {False}
\end{algorithmic}
\end{algorithm}

In response to our hypothesis, we formulate our approach step by step, as shown in Algorithm~\ref{GECScore_based_Detection}.

\paragraph{Step 1} Similar to previous zero-shot methods, \textsc{GECScore} requires a preliminary sample set $X = \{x_1, x_2, \ldots, x_{i-1}\}$ to help calculate the threshold. Given the sample $x_i$ to be detected, the entire sample set is expanded to ${X} = \{x_1, x_2, \ldots, x_{i-1}, x_i\}$, where \textit{i} represents the total number of samples used to construct the threshold.

\paragraph{Step 2} We first employ a grammar error correction function $g$, typically instantiated through a GEC model with an autoregressive or seq2seq architecture, to generate grammatically corrected results $\tilde{X} = \{\tilde{x_1}, \tilde{x_2}, \ldots, \tilde{x_{i-1}}, \tilde{x_i}\}$.

\paragraph{Step 3} We then use a similarity scoring metric $\mathrm{Sim}$ to calculate the similarity scores ${S} = \{{s}_1, {s}_2, ..., {s}_{i-1}, {s}_{i}\}$ between the grammatical corrected texts $\tilde{X}$ and the original texts $X$. 
In our study, we use ROUGE2~\cite{DBLP:conf/acl/SellamDP20}, which achieves the best average performance (see \autoref{table:rouge_metrics_performance}), as an example for scoring, and discuss other similarity scoring metrics in Appendix \ref{sec:similarity_metrics}. 

\paragraph{Step 4} Based on the calculation of the similarity scores ${S}$ for the entire sample set, determine the optimal threshold $\epsilon$ for detection. This threshold should achieve the best balance between the true positive rate (TPR) and the false positive rate (FPR).
The more diverse the sample types in the preliminary sample set, the better and more stable the performance of \textsc{GECScore}.

\paragraph{Step 5} Finally, when comparing the input sample $x_i$ with the threshold $\epsilon$, if the score of $x_i$ is greater than $\epsilon$, the sample $x_i$ is more likely generated by LLMs; otherwise, it is probably written by humans.

\begin{table*}[!ht]
\centering
\renewcommand{\arraystretch}{1.1}
\resizebox{\textwidth}{!}{
\begin{tabular}{ll cccccccccccc}
\toprule
\multirow{2}{*}{\bf Dataset$\downarrow$} & \multirow{2}{*}{\bf Method$\downarrow$} 
& \multicolumn{2}{c}{\textbf{GPT-3.5}} & \multicolumn{2}{c}{\textbf{PaLM2}} & \multicolumn{2}{c}{\textbf{GPT-4o}} & \multicolumn{2}{c}{\textbf{Claude-3.5}} & \multicolumn{2}{c}{\textbf{Llama-3}}  & \multicolumn{2}{c}{\textbf{Avg.}}\\ 
  &  & AUROC & $F_{1}$ & AUROC & $F_{1}$ & AUROC & $F_{1}$ & AUROC & $F_{1}$ & AUROC & $F_{1}$ & AUROC & $F_{1}$ \\ 
\midrule
\multirow{11}{*}{XSum} 
& Log-Likelihood & 91.39 & 86.99 & 93.05 & 88.73 & 78.93 & 70.10 & 93.34 & 89.45 & 97.25 & 97.31 & 90.79 & 86.92 \\
& Rank & 76.49 & 67.24 & 72.95 & 64.81 & 61.00 & 51.38 & 75.96 & 67.25 & 86.35 & 76.95 & 74.55 & 65.13 \\
& Log-Rank & 90.34 & 85.35 & 92.00 & 87.61 & 76.88 & 70.28 & 93.21 & 89.55 & 97.57 & 97.62 & 90.00 & 86.48 \\
& LRR & 81.04 & 72.82 & 80.90 & 76.45 & 66.88 & 56.54 & 91.11 & 84.98 & \textbf{99.55} & \textbf{97.97} & 83.90 & 77.35 \\
& NPR & 69.07 & 55.98 & 55.38 & 42.30 & 56.24 & 43.57 & 69.03 & 59.67 & 74.60 & 67.41 & 64.86 & 53.39 \\
& DetectGPT & 40.17 & 02.34 & 43.08 & 18.00 & 56.75 & 47.71 & 62.00 & 59.81 & 60.41 & 52.31 & 52.48 & 36.43 \\
& Fast-DetectGPT & 75.85 & 73.64 & 80.81 & 76.40 & 76.51 & 70.30 & 91.91 & 87.70 & 97.56 & 96.34 & 84.53 & 80.48 \\
& Revise-Detect & 94.98 & 91.47 & 79.78 & 74.91 & 76.81 & 73.63 & 81.24 & 79.06 & 93.04 & 88.77 & 85.17 & 81.97 \\
\cdashline{2-14}
& RoB-base & 61.86 & 53.41 & 68.12 & 50.81 & 56.87 & 59.81 & 65.93 & 57.00 & 96.97 & 93.94 & 69.95 & 63.79 \\
& RoB-large & 57.21 & 43.63 & 65.01 & 50.99 & 40.88 & 00.00 & 47.40 & 16.89 & 92.64 & 87.67 & 60.63 & 39.04 \\
\cdashline{2-14}
& \textsc{GECScore} (GPT-4o-Mini) & \textbf{99.25} & \textbf{95.69} & \textbf{94.31} & \textbf{92.29} & \textbf{98.97} & \textbf{95.04} & \textbf{99.14} & \textbf{96.55} & 98.38 & 94.29 & \textbf{98.01} & \textbf{94.77} \\
& \textsc{GECScore} (COEDIT-L) & 98.55 & 95.00 & 92.30 & 86.42 & 89.87 & 83.35 & 97.37 & 92.30 & 95.79 & 89.15 & 94.78 & 89.24 \\
\midrule
\multirow{11}{*}{WP}
& Log-Likelihood & 97.18 & 93.62 & 97.01 & 92.15 & 94.96 & 89.56 & 97.58 & 93.65 & 99.55 & 98.40 & 97.26 & 93.48 \\
& Rank & 94.04 & 86.65 & 91.01 & 83.28 & 86.74 & 79.44 & 89.62 & 82.23 & 94.87 & 88.21 & 91.26 & 83.96 \\
& Log-Rank & 96.51 & 91.61 & 96.35 & 90.03 & 94.17 & 88.06 & 97.23 & 92.40 & 99.71 & 98.60 & 96.79 & 92.14 \\
& LRR & 89.68 & 81.80 & 90.27 & 82.27& 88.21 & 80.07 & 93.13 & 85.04 & 99.52 & 97.60 & 92.16 & 85.36 \\
& NPR & 89.82 & 62.33 & 87.74 & 73.89 & 84.36 & 75.92 & 87.32 & 76.35 & 91.31 & 84.34 & 88.11 & 74.97 \\
& DetectGPT & 83.35 & 62.06 & 78.01 & 70.11 & 89.47 & 87.29 & 85.55 & 83.95 & 92.48 & 87.96 & 85.77 & 78.67 \\
& Fast-DetectGPT & \textbf{99.58} & 97.57 & 99.42 & 97.58 & \textbf{99.32} & 96.75 & 97.58 & 94.57 & \textbf{99.85} & \textbf{99.60} & 99.15 & 97.21 \\
& Revise-Detect & 98.97 & 96.46 & 92.69 & 90.65 & 81.23 & 75.65 & 58.60 & 48.48 & 86.25 & 82.36 & 83.55 & 78.32 \\
\cdashline{2-14}
& RoB-base & 59.08 & 45.55 & 70.30 & 57.32 & 60.06 & 51.44 & 57.29 & 46.41 & 96.08 & 91.95 & 68.56 & 58.93 \\
& RoB-large & 41.54 & 67.68 & 65.67 & 62.94 & 30.91 & 66.75 & 30.41 & 66.62 & 77.64 & 67.43 & 49.23 & 66.68 \\
\cdashline{2-14}
& \textsc{GECScore} (GPT-4o-Mini) & 99.23 & 98.07 & \textbf{99.44} & \textbf{98.07} & 99.18 & \textbf{97.46} & 98.39 & \textbf{95.53} & 99.49 & 97.37 & \textbf{99.23} & \textbf{97.30} \\
& \textsc{GECScore} (COEDIT-L) & 98.92 & \textbf{98.08} & 97.28 & 97.10 & 96.02 & 90.66 & \textbf{98.98} & 93.27 & 99.07 & 96.52 & 98.05 & 95.53 \\
\bottomrule
\end{tabular}
}
\caption{Baseline Comparisons across Datasets and Generative Models. Evaluates 10 baselines using AUROC and $F_{1}$-Score (\%) on 2 datasets and 5 LLMs with 500 LLM-generated and 500 human-written samples per dataset.}
\label{table:normal_comparation}
\end{table*}
\section{Experiment}
\subsection{Settings}

\paragraph{Task Setup}


We define our task as \textit{black-box zero-shot LLM-generated text detection}, combining two critical aspects: \textit{zero-shot} and \textit{black-box} detection. \textit{Zero-shot} refers to detecting LLM-generated text without training on paired human-written and LLM-generated samples, instead relying on inherent feature differences that objectively exist between the two types of text. This allows thresholds derived from one dataset or model (e.g., GPT) to generalize effectively to others (e.g., Claude) with minimal performance degradation. \textit{Black-box} detection reflects real-world scenarios where the source model is unknown. In such cases, traditional zero-shot methods may rely on a surrogate model to perform detection. In our study, we adopt this zero-shot approach under the black-box setting and, following \citet{DBLP:journals/corr/abs-2310-05130}, use GPT-Neo-2.7B\footnote{\url{https://huggingface.co/EleutherAI/gpt-neo-2.7B}} as the surrogate model in our experiments.

\paragraph{Datasets}

Following the methodology of DetectGPT~\cite{DBLP:conf/icml/Mitchell0KMF23}, we curated a dataset of human-written texts from various daily domains and typical LLM use cases, excluding SQuAD \cite{DBLP:conf/emnlp/RajpurkarZLL16} for the brevity of its text samples. Specifically, we selected XSum \cite{DBLP:conf/emnlp/NarayanCL18} to represent news writing, which typically exhibits a concise and formal writing style, and Writing Prompts \cite{DBLP:conf/acl/LewisDF18} to represent creative writing, which typically features less structured narratives and a less formal style. We extracted a balanced corpus comprising 500 samples from each domain, maintaining a minimum threshold of 300 words per sample to ensure reliable detection and substantial content for analysis. 
We use advanced generative models including GPT-3.5-
turbo~\cite{blog2022chatgpt}, PaLM-2-bison~\cite{blog2023palm}, GPT-4o~\cite{blog2024gpt4o}, Claude-3.5-Sonnet~\cite{blog2024claude3.5}, and Llama-3-70B~\cite{blog2024llama3} to verify the effectiveness of our proposed method. Comprehensive details of our dataset are listed in Appendix \ref{sec:dataset_settings}.

\paragraph{Baselines} 

We evaluated our method against a comprehensive suite of zero-shot detectors, including Log-Likelihood \cite{DBLP:journals/corr/abs-1908-09203}, Rank \cite{DBLP:conf/acl/GehrmannSR19}, Log-Rank \cite{DBLP:conf/acl/GehrmannSR19}, LRR \cite{DBLP:conf/emnlp/SuZ0N23}, NPR \cite{DBLP:conf/emnlp/SuZ0N23}, DetectGPT \cite{DBLP:conf/icml/Mitchell0KMF23}, Fast-DetectGPT \cite{DBLP:journals/corr/abs-2310-05130} and Revise-Detect \cite{DBLP:conf/emnlp/ZhuYCCFHDL0023}. For supervised detectors, we utilized the OpenAI suite\footnote{\url{https://github.com/openai/gpt-2-output-dataset/tree/master/detector}} of detectors, which were fine-tuned on PLMs of RoBERTa-base and RoBERTa-large architecture \cite{DBLP:journals/corr/abs-1907-11692}.

\paragraph{GEC Model} The models we employed for performing grammar error correction include:

\begin{itemize}
    \item \textbf{Generative LLMs}: We use GPT-4o-Mini~\cite{blog2024gpt4o} as the main GEC model for the experiment, which is a lightweight and cost-efficient model that ensures effective detection tasks at low cost and low latency. This model exhibits enhanced natural language understanding and reasoning capabilities, fulfills the requirements for grammatical and stylistic modifications, and supports multiple languages \cite{DBLP:journals/corr/abs-2304-01746}. In our setup, the GEC prompt template used was ``Correct the grammar errors in the following text: <text for detection>\textbackslash nCorrected text:''.

    \item \textbf{Seq2seq LLMs}: We also use COEDIT-L~\cite{DBLP:conf/emnlp/Raheja0KK23} as a low-cost alternative GEC model for our method, providing an approach without API overhead and faster than generative LLMs. The model is built on the Google's Flan-T5-Large~\cite{DBLP:journals/corr/abs-2210-11416} architecture, achieving SOTA results in various text editing benchmarks.
\end{itemize}

\paragraph{Metrics} 

We used the AUROC and $F_{1}$-Score for evaluation. Each experiment included an equal number of samples from both LLM-generated and human-written texts to ensure balance.
\subsection{Main Results}

The experimental results in \autoref{table:normal_comparation} demonstrate that \textsc{GECScore} consistently achieves the highest average AUROC on the XSum and Writing Prompts datasets. Notably, when GPT-4o-Mini is used as the GEC model, it exhibits remarkable capability in detecting LLM-generated text. Specifically, \textsc{GECScore} (GPT-4o-Mini) delivers stable and outstanding performance across all datasets, with an average AUROC of 98.6\%. Compared to the SOTA detector Fast-DetectGPT, \textsc{GECScore} achieves an average improvement of 6.78\% in AUROC and 7.19\% in $F_{1}$-Score. Additionally, \textsc{GECScore} (GPT-4o-Mini) surpasses all supervised detectors, achieving a 29.36\% improvement in AUROC and a 34.67\% increase in $F_{1}$-Score.
Furthermore, although the XSum dataset poses greater challenges compared to Writing Prompts, \textsc{GECScore} (GPT-4o-Mini) achieves the best AUROC across most LLM settings, ranking second only to LRR in the case of Llama-3. This highlights its superior detection capability for formal writing texts.
On the Writing Prompts dataset, while Fast-DetectGPT proves to be a strong competitor, \textsc{GECScore} (GPT-4o-Mini) demonstrates comparable performance, further showcasing its robustness.

\subsection{Feature Contributions}

Our approach leverages two key features: \textbf{Differences in Grammatical Errors} and \textbf{Differences in LLM Correction Preferences}. Our baseline design validates the relevance of these two features. Overall, the inherent differences in grammatical errors between the two types of text serve as the primary factor for distinguishing between them. The experimental results of \textsc{GECScore} (COEDIT-L) provide strong evidence for this conclusion (see \autoref{table:normal_comparation}). Unlike GPT-4o, COEDIT-L is a seq2seq model specifically fine-tuned for the GEC task. However, it lacks the correction preferences exhibited by LLMs. The results show that \textsc{GECScore} (COEDIT-L) achieves an AUROC of 98.05\% on Writing Prompts and 94.78\% on XSum, outperforming the SOTA method Fast-DetectGPT.

The contribution of differences in LLM correction preferences was less significant than that of differences in grammatical errors. Specifically, we compared the results of our method with Revise-Detect~\cite{DBLP:conf/emnlp/ZhuYCCFHDL0023}, a method that relies solely on LLM correction preference differences (i.e., the consistency of LLM outputs before and after revision). As shown in \autoref{table:normal_comparation}, the AUROC of Revise-Detect is approximately 15\% lower than that of \textsc{GECScore} (COEDIT-L). Nonetheless, differences in LLM correction preferences enhance the performance and robustness of our approach. For example, on the XSum dataset, the average AUROC of \textsc{GECScore} (GPT-4o-Mini) is 98.01\%, while the average AUROC of \textsc{GECScore} (COEDIT-L) is 94.78\%. This additional performance improvement of GPT-4o-Mini can be attributed to the correction preferences of the LLM.

In summary, we found that explicitly guiding the LLM to perform specific feature-related tasks (e.g., grammatical error correction) can more effectively extract the feature differences between two categories of text. This approach can improve detection performance compared to a simple rewriting task.

\subsection{Ablation Study}

\begin{figure*}[!ht]
    \centering
    \includegraphics[width=\textwidth, trim=0 0 0 0]{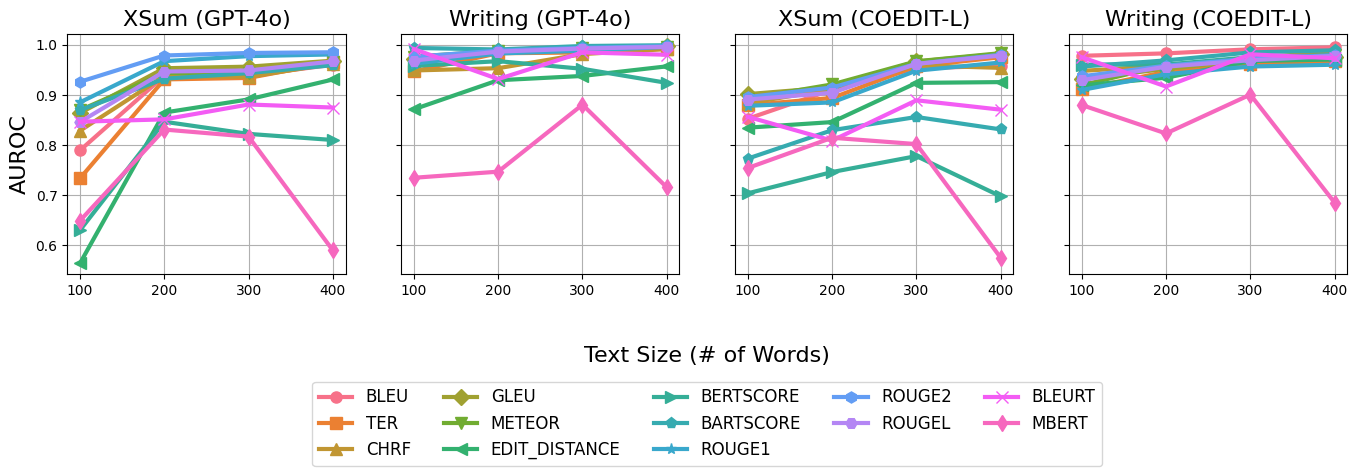}
    \caption{Impact of Different GEC Models and Scoring Metrics Performance on Different Text Sizes. The plot displays the AUROC across varying text sizes. The x-axis represents the number of words of the text, while the y-axis indicates the corresponding detection performance of \textsc{GECScore} with different settings.}
    \label{fig:ablation_analysis}
\end{figure*}

We investigated the factors influencing the efficacy of \textsc{GECScore}, focusing on the scoring metric, GEC model, and text size. The experimental settings are described in \autoref{ablation_settings}. Figure \ref{fig:ablation_analysis} presents the results of our empirical analysis. Scoring metrics are identified as a critical factor affecting the effectiveness of \textsc{GECScore}.

\subsubsection{Impact of Scoring Metric}
\begin{table*}[!ht]
\centering
\renewcommand{\arraystretch}{1.2} 
\resizebox{\textwidth}{!}{
\begin{tabular}{l | cccccccccccc}
\toprule
\textbf{Metrics $\rightarrow$} & \textbf{BERTScore} & \textbf{Edit Distance} & \textbf{BLEURT} & \textbf{TER} & \textbf{chrF} & \textbf{BLEU} & \textbf{ROUGEL} & \textbf{METEOR} & \textbf{BARTScore} & \textbf{GLEU} & \textbf{ROUGE1} & \textbf{ROUGE2} \\ 
\midrule
GPT-4o-mini & 86.39 & 86.84 & 91.79 & 93.41 & 94.27 & 94.65 & 95.67 & 95.97 & 96.10 & 96.17 & 96.77 & \textbf{97.89} \\
\midrule
COEDIT-L & 85.03 & 91.54 & 90.91 & 93.67 & 94.69 & 95.28 & 94.63 & 94.90 & 89.88 & 94.95 & 93.08 & 95.01 \\
\bottomrule
\end{tabular}
} 
\caption{Average AUROC (\%) of the performance of the difference scoring metrics and scoring metrics for different text sizes. ROUGE2 achieves the best performance among all scoring metrics for \textsc{GECScore} (GPT-4o-mini) and the second best performance among all scoring metrics for \textsc{GECScore} (COEDIT-L), which we use as an example in this paper.}
\label{table:rouge_metrics_performance}
\end{table*}

Metrics that integrate semantic understanding and analyze language units at a granular level often yield better results in our approach. Specifically, ROUGE2~\cite{lin2004rouge} achieves optimal performance (see \autoref{table:rouge_metrics_performance}) when GPT-4o-Mini is used as the GEC model, with an average AUROC of 97.89\%. It also ranks second when COEDIT-L is employed. Additionally, metrics that consider longer sequences and semantic changes, such as ROUGEL~\cite{lin2004rouge}, BARTScore~\cite{yuan2021bartscore}, and METEOR~\cite{BanerjeeL05}, exhibit strong performance. However, not all semantic-based metrics perform equally well. The effectiveness of such metrics heavily depends on task-specific fine-tuning. While neural-based metrics like BERTScore~\cite{zhangbertscore} and BLEURT~\cite{DBLP:conf/acl/SellamDP20} show potential, their poor task-specific adaptation makes them less sensitive to subtle linguistic changes. This limitation leads to weaker performance, with BERTScore achieving only 85.71\% average AUROC and showing inconsistent stability across text sizes.

Metrics that evaluate overall structure and lexical accuracy also perform strongly. These metrics, such as TER~\cite{SnoverDSMM06}, chrF~\cite{Popovic15}, and BLEU~\cite{papineni2002bleu}, achieve average performances of approximately 93.54\%, 94.48\%, and 94.96\%, respectively, across both \textsc{GECScore} employing GPT-4o-Mini and COEDIT-L. Their detection abilities improve with text size, maintaining consistent performance. In contrast, edit-based metrics like Edit Distance~\cite{Navarro01} focus more on surface-level changes, leading to greater variability in performance.

\subsubsection{Impact of GEC Model}

Generative LLMs (GPT-4o-Mini) demonstrate stronger average performance as the GEC model for our method. While both GPT-4o-Mini and COEDIT-L generally perform consistently,  GPT-4o-Mini excels on scoring metrics like ROUGE1, ROUGE2, GLEU, and METEOR. They also show stronger detection capabilities on samples with text sizes ranging from 150 to 300 words. However, for texts of approximately 100 words, \textsc{GECScore} (GPT-4o-Mini), based on metrics that evaluate structure and lexical accuracy, such as BLEU and TER, exhibits weaker performance, achieving only about 80\% AUROC. In contrast, it performs well using metrics based on semantic understanding and sequence transformation, such as ROUGE and METEOR. For shorter texts or when lower overhead is needed, the Seq2seq (COEDIT-L) model serves as a suitable alternative.

\subsubsection{Impact of Text Size}

As text size increases, most metrics show either improvement or stabilization in AUROC. \textsc{GECScore} generally performs better with longer texts, suggesting that additional information in longer texts enhances evaluation accuracy. When using GPT-4o-Mini as the GEC model, \textsc{GECScore} typically maintains consistent performance across various text sizes. However, for shorter texts of approximately 100 words, there is slight variation in performance among metrics.
Despite this, \textsc{GECScore} with GPT-4o-Mini remains robust in the optimal metric, ROUGE2, maintaining over 98\% AUROC for texts exceeding 200 words. In contrast, \textsc{GECScore} with COEDIT-L shows greater variability, struggling to achieve strong performance on XSum until the text size exceeds 300 words.
While most metrics improve with larger text sizes, \textsc{GECScore} using metrics such as mBERTScore, BERTScore, and BLEURT may face limitations when evaluating longer texts.

\subsection{Reliability in the Wild}

\subsubsection{Varied Text Sources}

Our detector is designed to achieve high generalizability and usability across various domains and generative models. To evaluate its effectiveness, we compared its performance with other detection techniques across diverse datasets and models. As shown in \autoref{fig:ablation_analysis}, our method consistently outperforms others on the XSum and Writing Prompts. In contrast, zero-shot methods struggle more with XSum than with Writing Prompts, likely due to the distinct stylistic features of these domains.

Our method stands out by maintaining impressive performance on texts generated by various models, with an average AUROC of 96\%. Notably, \textsc{GECScore} (GPT-4o-Mini) achieves an AUROC of more than 94\% and an $F_{1}$-score of 92\% across all evaluation settings. Our analysis reveals that the type of generative model further amplifies the impact of domain type on current zero-shot methods. For example, on the XSum dataset, Fast-DetectGPT \cite{DBLP:journals/corr/abs-2310-05130} achieves over 90\% AUROC on data generated by Claude-3.5 and Llama-3, but only around 80\% AUROC on data generated by GPT-3.5, PaLM2, and GPT-4o. 
Additionally, the black-box zero-shot detection method Revise-Detect \cite{DBLP:conf/emnlp/ZhuYCCFHDL0023} demonstrates competitive performance on text generated by GPT-3.5-turbo, achieving an AUROC score of 96.97\%. However, its performance on text generated by other LLMs, such as PaLM2, is less consistent, achieving only 79.78\% AUROC on XSum but reaches up to 92.69\% on Writing Prompts, with an average AUROC of only 80.14\% across all settings. In contrast, \textsc{GECScore} maintains consistent and reliable performance across domains and generators. This highlights its potential to remain competitive even in other unknown domains or generators.

\subsubsection{Generalization on Text Source}
\label{sec:generalization}
\paragraph{Cross-Domain}

We evaluated the generalization ability of \textsc{GECScore} across the XSum and Writing Prompt datasets. For these two domains, we used one as training data to determine the decision threshold and tested its performance on the other dataset. The results in \autoref{table:cross_domain} show that \textsc{GECScore} (GPT-4o-Mini) demonstrates better generalization ability in cross-domain scenarios compared to other baseline methods. For example, when using the threshold trained on XSum to detect Writing Prompt and using the threshold trained on Writing Prompt to detect XSum, the performance of \textsc{GECScore} (GPT-4o-Mini) decreased by an average of only 3.39\% in terms of AUROC. In contrast, Fast-DetectGPT's performance decreased by an average of 8.12\% in terms of AUROC. Furthermore, \textsc{GECScore} (GPT-4o-Mini) consistently maintains optimal performance, staying above 94\% AUROC, demonstrating strong robustness.

\begin{table}[!ht]
\centering
\renewcommand{\arraystretch}{1.2}
\resizebox{0.5\textwidth}{!}{
\begin{tabular}{l | cccccccccccc}
\toprule
\textbf{Dataset} & \textbf{XSum $\rightarrow$ WP} \ & \textbf{WP $\rightarrow$ XSum} \\ 
\midrule
Log-Likelihood & 91.70 ($\downarrow$05.48) & 76.00 ($\downarrow$15.39) \\
Rank & 84.60 ($\downarrow$09.44) & 68.90 ($\downarrow$07.59) \\
Log-Rank & 88.99 ($\downarrow$07.52) & 70.50 ($\downarrow$19.84) \\
LRR & 79.40 ($\downarrow$10.28) & 71.00 ($\downarrow$10.04 )\\
NPR & 79.21 ($\downarrow$10.61) & 63.20 ($\downarrow$05.87) \\
DetectGPT & 51.27 ($\downarrow$32.08) & 48.58 ($\downarrow$00.00) \\
Fast-DetectGPT & 93.19 ($\downarrow$06.39) & 66.00 ($\downarrow$09.85) \\
Revise-Detect & 91.00 ($\downarrow$07.97) & 88.40 ($\downarrow$06.58) \\
Rob-base & 56.60 ($\downarrow$02.48) & 59.10 ($\downarrow$02.76) \\
\textsc{GECScore} (GPT-4o-Mini) & \textbf{97.60} ($\downarrow$01.63) & \textbf{94.10} ($\downarrow$05.15) \\
\textsc{GECScore} (COEDIT-L) & 89.50 ($\downarrow$09.42) & 80.29 ($\downarrow$18.26) \\
\bottomrule
\end{tabular}
}
\caption{Results of Cross-Domain in terms of AUROC (\%). The
values in parentheses represent the performance decrease compared to in-distribution testing.}
\label{table:cross_domain}
\end{table}

\paragraph{Cross-Generator}

We generalize \textsc{GECScore} between datasets generated by GPT-3.5 and Claude-3.5 in a similar setting to Cross-Domain. As shown in \autoref{table:cross_generator}, \textsc{GECScore} (GPT-4o-Mini) also has better generalization capabilities than other baselines on Cross-Generator, with performance dropping by an average of only 3.6\% AUROC. The average performance remains at over 95\% AUROC, exhibiting a trend similar to that observed in Cross-Domain scenarios. Notably, the average performance of \textsc{GECScore} (COEDIT-L) drops by only 4.71\% AUROC in Cross-Generator scenarios but decreases significantly by 13.84\% AUROC in Cross-Domain scenarios. This indicates its limitation in Cross-Domain, while maintaining acceptable performance in Cross-Generator scenarios.

\begin{table}[!ht]
\centering
\renewcommand{\arraystretch}{1.2}
\resizebox{0.5\textwidth}{!}{
\begin{tabular}{l | cccccccccccc}
\toprule
\textbf{Dataset} & \textbf{GPT $\rightarrow$ Claude} & \textbf{Claude $\rightarrow$ GPT} \\
\midrule
Log-Likelihood & 87.30 ($\downarrow$06.04) & 85.40 ($\downarrow$05.99) \\
Rank & 70.20 ($\downarrow$05.76) & 69.30 ($\downarrow$07.19) \\
Log-Rank & 87.90 ($\downarrow$05.31) & 84.20 ($\downarrow$02.44) \\
LRR & 83.50 ($\downarrow$07.61) & 72.90 ($\downarrow$08.14 )\\
NPR & 65.65 ($\downarrow$03.38) & 63.65 ($\downarrow$05.42) \\
DetectGPT & 49.41 ($\downarrow$12.59) & 44.37 ($\downarrow$00.00) \\
Fast-DetectGPT & 79.10 ($\downarrow$12.81) & 67.20 ($\downarrow$08.65) \\
Revise-Detect & 66.48 ($\downarrow$14.76) & 75.50 ($\downarrow$19.48) \\
Rob-base & 63.00 ($\downarrow$02.93) & 58.60 ($\downarrow$03.26) \\
\textsc{GECScore} (GPT-4o-Mini) & \textbf{95.59} ($\downarrow$03.55) & \textbf{95.60} ($\downarrow$03.65) \\
\textsc{GECScore} (COEDIT-L) & 91.80 ($\downarrow$05.57) & 94.70 ($\downarrow$03.85) \\
\bottomrule
\end{tabular}
}
\caption{Results of Cross-Generator in terms of AUROC (\%). The values in parentheses represent the performance decrease compared to in-distribution test.}
\label{table:cross_generator}
\end{table}

\subsubsection{Paraphrase Attack}

In practical applications, LLM-generated text detectors may be exposed to attacks simulating human post-editing of LLM-generated text \cite{DBLP:journals/corr/abs-2310-14724}. To evaluate the effectiveness and robustness of our approach, we follow the approach of \citet{DBLP:journals/corr/abs-2303-11156} and use a T5 paraphrase model\footnote{\url{https://huggingface.co/Vamsi/T5_Paraphrase_Paws}} to perform the paraphrase attack on our detector. The results presented in \autoref{table:paraphrased} indicate that the performance of all baseline methods decreases significantly when faced with paraphrased text. Specifically, all zero-shot methods experience an average decrease in performance of 12.93\% AUROC when encountering paraphrase attacks. In contrast, our method achieves the most competitive detection performance in the paraphrase attack scenario, demonstrating superior detection capability.
\section{Conclusion}

In this paper, we propose the hypothesis that, from the perspective of LLMs, human-written texts typically contain more grammatical errors than LLM-generated texts, and we verify this through a statistical analysis experiment. To improve the performance of detectors in real-world black-box scenarios, we introduce a simple yet effective zero-shot detection method called \textsc{GECScore}. \textsc{GECScore} is a black-box tool that can effectively identify text generated by LLMs without access to the source model or a large amount of training data. Experimental results demonstrate that \textsc{GECScore} surpasses the current SOTA detectors, including both supervised and zero-shot methods. Furthermore, \textsc{GECScore} exhibits strong reliability in various real-world scenarios, showing robust performance across texts from different sources, strong generalization ability, and resistance to paraphrase attacks.

\section*{Limitations}

\begin{itemize}
    \item \textbf{Textual Integrity} For \textsc{GECScore} to function effectively, it is crucial that the text samples being analyzed are complete. Fragmented sentences can be inadvertently completed by the GEC model, which may disrupt and confuse \textsc{GECScore}'s detection mechanism.
    \item \textbf{Grammar Attacks} A noteworthy potential attack on \textsc{GECScore} involves injecting grammatical errors into the samples. This could significantly increase the chances of LLM-generated text being mistaken for human-written text. However, this approach may not be practical in real-world scenarios, as introducing grammatical errors would degrade the quality of the text and damage the author's reputation. Furthermore, we do not consider texts that have been grammatically corrected by the LLM, as such texts cannot be strictly defined as human-written.
\end{itemize}

\section*{Acknowledgments}

This work was supported in part by the Science and Technology Development Fund of Macau SAR (Grant No. 0007/2024/AKP), the Science and Technology Development Fund of Macau SAR (Grant No. FDCT/0070/2022/AMJ, China Strategic Scientific and Technological Innovation Cooperation Project Grant No. 2022YFE0204900), the Science and Technology Development Fund of Macau SAR (Grant No. FDCT/060/2022/AFJ, National Natural Science Foundation of China Grant No. 62261160648), the UM and UMDF (Grant Nos. MYRG-GRG2023-00006-FST-UMDF, MYRG-GRG2024-00165-FST-UMDF), and the National Natural Science Foundation of China (Grant No. 62266013). This work was performed in part at SICC which is supported by SKL-IOTSC, and HPCC supported by ICTO of the University of Macau. We thank all anonymous reviewers for their insightful comments.

\bibliography{custom}

\appendix

\section{Experimental Settings}
\subsection{Black-box Zero-shot Detection Task}

Our research focuses on black-box zero-shot LLM generated text detection, under the stipulation that our proposed method has not been trained on any human-written and LLM-generated text pairs or has access to the source model of generated text. 

\paragraph{Zero-shot Detection} 

Our zero-shot detection setup aligns with previous mainstream works such as DetectGPT~\cite{DBLP:conf/icml/Mitchell0KMF23} and Fast-DetectGPT~\cite{DBLP:journals/corr/abs-2310-05130}, where specific features are extracted from a sample set to distinguish between human-written texts and those generated by LLMs. The term ``zero-shot'' indicates that the differences in the features used objectively exist between the two types of texts. Therefore, by determining a predefined feature classification threshold, it is possible to detect generated texts from any LLMs or domains. As shown in Section~\ref{sec:generalization}, the threshold extracted from the XSum dataset can be effectively applied to the Writing Prompts dataset, while the threshold obtained from GPT-generated texts works seamlessly on the Claude-generated texts, with almost no performance degradation.

\paragraph{Black-box Detection} 

This research focus is particularly consistent with real-world scenarios, where the specific identity of the generative model employed for text generation often remains obscured. This means that identifying LLM-generated text from an unknown model is increasingly important, and we refer to this task as the ``Black-box Detection Task.''

The setting of the ``Black-box Detection Task'' significantly distinguishes our approach from both traditional zero-shot detection methods and supervised fine-tuning classifiers. Traditional zero-shot detection methods require access to the language model to identify text token irregularities. However, in the ``Black-box Detection Task,'' the traditional zero-shot method must rely on a surrogate model instead of the source model to perform the detection task. In contrast, supervised classifiers that are fine-tuned typically harness language models to distill vectorized representations of textual features, a process contingent on a substantial corpus of annotated samples for training purposes.

Our zero-shot methodology diverges from these traditional strategies by incorporating an auxiliary grammatical error correction (GEC) model, which accentuates and leverages the grammatical discrepancies between human-written text and text generated by LLMs. In this study, we primarily conduct experiments under the ``Black-box Detection Task'' settings. Following the experiments of \citet{DBLP:journals/corr/abs-2310-05130}, we use GPT-Neo-2.7B\footnote{\url{https://huggingface.co/EleutherAI/gpt-neo-2.7B}} as the surrogate model for the traditional zero-shot method.

\subsection{Hypothesis Validation Settings}
\label{sec:hypothesis_validation_settings}

To validate our hypothesis, we conducted a comparative analysis of grammatical errors in human-written texts and those generated by LLMs. Due to the significant resource demands of manual annotation, we opted to use GPT-4o for identifying and annotating errors instead of relying on human annotators. Our experimental dataset includes human-written samples from Writing Prompts and texts generated by GPT-3.5-turbo~\cite{blog2022chatgpt}, PaLM-2-bison~\cite{blog2023palm}, Claude-3.5-Sonnet~\cite{blog2024claude3.5}, and Llama-3-70B~\cite{blog2024llama3}. We set the \texttt{temperature} parameter of GPT-4o~\cite{blog2024gpt4o} to \texttt{0.01} and \texttt{max\_tokens} to \texttt{1000}. The prompt used for identifying and counting grammatical errors is as follows:

\begin{lstlisting}[language=bash, caption=Message for identifying and counting grammatical errors]
[
    {'role': 'user', 'content': 'Text1 is the original text, Text2 is the grammatical error corrected version of Text1. Please analyze how many grammatical errors were corrected in Text2 compared to Text1, and list them. Then, categorize the statistics according to the three editing types: addition, deletion, and modification. Please end the result with "Total: (a number), Addition: (a number), Deletion: (a number), Substitution: (a number)".
    
    Text1: <orignal_text>
    Text2: <grammatical_corrected_text>'}
]
\end{lstlisting}


In the message context, <original\_text> refers to the original text, and <grammatical\_corrected\_text> refers to the text after grammar error correction.

\subsection{Datasets Settings}
\label{sec:dataset_settings}

\paragraph{Human Datasets} 

We follow \citet{DBLP:conf/icml/Mitchell0KMF23} and select human-written texts covering various everyday domains and practical LLM application cases to construct an LLM-generated text dataset. Specifically, XSum~\cite{DBLP:conf/emnlp/NarayanCL18} is used to represent news writing, and Writing Prompts~\cite{DBLP:conf/acl/LewisDF18} is used to represent creative writing. We did not use SQuAD~\cite{DBLP:conf/emnlp/RajpurkarZLL16} because its text samples are too short.
For our experiments, we extracted a balanced set of 500 human-written samples and 500 LLM-generated samples for the detection task. This was done to ensure a robust and rigorous analysis. Furthermore, we imposed a stringent criterion for the length of the text samples: each sample in our study was required to contain a minimum of 300 words to facilitate a more accurate assessment of our approach's reliability.

\paragraph{Generative Models and Settings}

\begin{table*}[!ht]
\centering
\small
\renewcommand{\arraystretch}{1.1}
\resizebox{0.8\textwidth}{!}{
\begin{tabular}{l l l}
\toprule
\textbf{Generative Model} &
\textbf{Model Path / Service} &
\textbf{Parameters} \\ 
\midrule

GPT-3.5-
turbo~\cite{blog2022chatgpt} & OpenAI/gpt-3.5-turbo & N/A \\
PaLM-2-bison~\cite{blog2023palm} & Google/chat-bison@002 & N/A \\
GPT-4o~\cite{blog2024gpt4o} & OpenAI/gpt-4o & N/A \\
Claude-3.5-Sonnet~\cite{blog2024claude3.5} & Anthropic/Claude-3.5-Sonnet & N/A \\
Llama-3-70B~\cite{blog2024llama3} & meta-llama/Meta-Llama-3-70B & 70B \\
\bottomrule

\end{tabular}
}
\caption{Details of the source models that is used to generate LLM-generated text.}
\label{table:generative_models}
\label{tab:llms}
\end{table*}

We use advanced generative models including GPT-3.5-
turbo~\cite{blog2022chatgpt}, PaLM-2-bison~\cite{blog2023palm}, GPT-4o~\cite{blog2024gpt4o}, Claude-3.5-Sonnet~\cite{blog2024claude3.5}, and Llama-3-70B~\cite{blog2024llama3} to verify the effectiveness of our proposed method. Comprehensive details of our dataset are listed in Appendix \ref{sec:dataset_settings}.

The temperature setting for text generation was maintained at the default value of 1 to foster the generation of creative content. Furthermore, the parameters \texttt{top\_p} and \texttt{top\_k} were configured to \texttt{40} and \texttt{0.96}, respectively.  We provide the first sentence of the text as a prefix to guide the model to perform the generation task. In order to ensure the length consistency of the LLM-generated text and human-written text, we set \texttt{min\_length} to \texttt{800} and \texttt{max\_length} to \texttt{1024}. This ensured the production of sufficiently lengthy texts, which were subsequently truncated to lengths closely mirroring those of human-written samples, all the while maintaining sentence integrity.

For closed-source models, we use API services to perform text generation tasks in the form of conversations. Referring to the experimental prompt of Fast-DetectGPT \cite{DBLP:journals/corr/abs-2310-05130}, we instructed the models to assume the personas of news and fiction writers to generate news articles and narrative stories respectively. The temperature of text generation also uses the default parameter 1. We provide the ``word length'' and ``first sentence text'' of the human-written sample in the prompt as <word count> and <prefix> to provide additional information to guide the LLMs generate a text that aligns with the specified requirements. The specific API Messages are as follows:

\begin{lstlisting}[language=bash, caption=Message for XSum]
[
    {'role': 'system', 'content': 'You are a News writer.'},
    {'role': 'user', 'content': 'Please write an article with about <word count> words starting exactly with: <prefix>'},
]
\end{lstlisting}

\begin{lstlisting}[language=bash, caption=Message for Writing Prompts]
[
    {'role': 'system', 'content': 'You are a Fiction writer.'},
    {'role': 'user', 'content': 'Please write an article with about <word count> words starting exactly with: <prefix>'},
]
\end{lstlisting}

\subsection{Details of the Compared Detection
Methods}

The details of the baseline methods we compared in our experiments are as follows:

\paragraph{Log-Likelihood} This method employs a language model to calculate the log probability for each token within a text \cite{DBLP:conf/acl/GehrmannSR19,DBLP:conf/icml/KirchenbauerGWK23}. Specifically, we calculate the average log probabilities of all tokens, thereby producing a score that reflects the text's Log-Likelihood. Higher scores indicate a greater probability of the text being LLM-generated.

\paragraph{Rank} In this approach, the absolute rank of each word within a text is determined based on its preceding context \cite{DBLP:conf/acl/GehrmannSR19}. A text's overall score is then calculated by averaging these rank values across all tokens. Similar to Log-Likelihood, lower scores suggest a higher probability of the text being LLM-generated.

\paragraph{Log-Rank} This method evaluates the contextual prominence of each token in a text using the Log-Rank metric \cite{DBLP:conf/acl/GehrmannSR19}. Unlike the traditional Rank metric, which relies on absolute word ranks, Log-Rank applies a logarithmic transformation to each token's rank value. The process involves computing the logarithm of rank positions for all tokens and aggregating these log-transformed ranks into a composite score.

\paragraph{LRR} \citet{DBLP:conf/emnlp/SuZ0N23} proposed the Log-Likelihood Log-Rank Ratio (LRR), an improved zero-shot method that combines Log-Likelihood and Log-Rank to enhance detection performance by providing complementary information. Texts with higher LRR scores are more likely to be generated by the target LLM.

\paragraph{NPR} The Normalized Perturbed Log-Rank (NPR) \cite{DBLP:conf/emnlp/SuZ0N23} applies perturbations to text to exploit differences in sensitivity between LLM-generated and human-written texts. LLM-generated texts typically show a more substantial increase in Log-Rank scores after perturbations, leading to higher NPR scores for LLM-generated texts compared to human-written texts.

\paragraph{DetectGPT} \citet{DBLP:conf/icml/Mitchell0KMF23} introduced DetectGPT, a method that evaluates changes in a model's log probability function in response to slight text modifications. The underlying theory posits that LLM-generated texts are often situated at local optima of the model's log probability function, resulting in a decrease in log probability for perturbed machine-generated texts. In contrast, similar perturbations applied to human-written texts may result in either an increase or decrease in log probability.

\paragraph{Fast-DetectGPT} While DetectGPT has demonstrated strong zero-shot detection capabilities, its high computational cost has limited its practical application. \citet{DBLP:journals/corr/abs-2310-05130} introduced Fast-DetectGPT, which leverages conditional probability curvature to highlight word choice differences between LLMs and humans. By replacing DetectGPT's perturbation technique with a more efficient sampling method, Fast-DetectGPT achieves a remarkable 340-fold increase in detection speed compared to DetectGPT, while maintaining high performance across various test scenarios.

\paragraph{Revise-Detect} \citet{DBLP:conf/emnlp/ZhuYCCFHDL0023} proposed a zero-shot, black-box method to detect LLM-generated texts by revising the target text with the ChatGPT model. The hypothesis is that ChatGPT will make fewer changes to LLM-generated texts compared to human-written texts, as the former align more closely with the generative patterns and statistical models used by LLMs. A higher similarity between the original text and its ChatGPT-modified version indicates a greater likelihood of the text being LLM-generated.

\paragraph{OpenAI Detector} \citet{DBLP:journals/corr/abs-1908-09203} released a dataset containing GPT-2 outputs and WebText samples to facilitate research on distinguishing human-written texts from LLM-generated ones. Using this dataset, they fine-tuned classifiers\footnote{\url{https://github.com/openai/gpt-2-output-dataset}} of RoBERTa-base and
RoBERTa-large architecture for LLM-generated text detection. 

\section{Robustness Against Paraphrase Attack}

\begin{table*}[]
\centering
\renewcommand{\arraystretch}{1.1}
\resizebox{\textwidth}{!}{
\begin{tabular}{l | llllll | lllllll}
\toprule
\textbf{Datasets $\rightarrow$} & \multicolumn{6}{c|}{\textbf{XSum}}   & \multicolumn{6}{c}{\textbf{Writing Prompts}} \\ 
\multicolumn{1}{l|}{\textbf{Method $\downarrow$}} & GPT-3.5 & PaLM2 & GPT-4o & Claude-3.5 & Llama-3 & \multicolumn{1}{c|}{\textbf{Avg.}} & GPT-3.5 & PaLM2 & GPT-4o & Claude-3.5 & Llama-3 & \textbf{Avg.} \\ 
\midrule
\multicolumn{13}{c}{\textit{Zero-shot Methods}} \\
\midrule
Log-Likelihood & 76.94 & \textbf{79.84} & 48.31 & 76.00 & 94.70 & 75.96 & 85.55 & 85.63 & 72.07 & 77.50 & \textbf{93.98} & 82.95 \\
Rank & 61.69 & 57.61 & 40.62 & 58.39 & 74.04 & 58.07 & 80.59 & 79.26 & 65.67 & 67.41 & 84.05 & 75.40 \\
Log-Rank & 76.56 & 78.82 & 48.06 & 76.89 & 95.46 & 75.96 & 84.05 & 84.15 & 71.59 & 76.92 & 94.87 & 82.72 \\
LRR & 72.32 & 74.07 & 49.41 & 76.94 & \textbf{97.31} & 74.01 & 77.50 & 77.32 & 69.21 & 73.98 & 95.77 & 78.76 \\
NPR & 55.32 & 45.60 & 39.04 & 53.61 & 61.98 & 51.51 & 72.65 & 76.14 & 65.03 & 68.34 & 80.73 & 72.18 \\
DetectGPT & 27.17 & 30.55 & 33.42 & 47.51 & 42.76 & 36.68 & 58.62 & 67.40 & 79.48 & 74.81 & 81.96 & 72.05 \\
Fast-DetectGPT & 47.90 & 65.70 & 43.79 & 69.14 & 91.75 & 63.26 & 92.52 & 93.82 & 80.26 & 74.72 & 97.71 & 87.81 \\
Revise-Detect & 91.23 & 79.43 & 72.17 & 82.71 & 87.48 & 82.20 & \textbf{96.72} & 94.29 & 77.11 & 65.41 & 86.01 & 83.91 \\
\midrule
\multicolumn{13}{c}{\textit{Supervised Methods}} \\
\midrule
RoB-base & 60.08 & 61.84 & 52.86 & 57.09 & 88.55 & 64.88 & 52.98 & 60.30 & 52.86 & 52.89 & 83.58 & 60.92 \\
RoB-large & 57.47 & 60.13 & 42.32 & 40.60 & 76.65 & 55.43 & 35.82 & 51.97 & 35.10 & 32.81 & 59.58 & 43.06 \\
\midrule
\multicolumn{13}{c}{\textit{Our Zero-shot Method: \textsc{GECScore}}} \\
\midrule
\textsc{GECScore} (GPT-4o-Mini) & \textbf{94.58} & 71.06 & \textbf{93.77} & \textbf{83.11} & 94.05 & \textbf{87.71} & 92.76 & 94.04 & 85.92 & 83.31 & 85.22 & 88.25 \\
\textsc{GECScore} (COEDIT-L) & 78.43 & 67.76 & 68.87 & 77.34 & 70.82 & 72.64 & 96.17 & \textbf{96.28} & \textbf{92.26} & \textbf{93.56} & 93.80 & \textbf{94.41} \\
\bottomrule
\end{tabular}
}
\caption{Comparison with 10 other baselines on 2 given datasets and 5 given generative models in paraphrased data settings. The evaluation metric is AUROC(\%). We evaluate using 500 LLM-generated samples and 500 human-written samples per dataset.}
\label{table:paraphrased}
\end{table*}

The robustness of the detection method to adversarial challenges is of paramount importance, particularly in practical applications where LLM-generated text detectors may face potential attacks, such as human post-editing of the LLM-generated text. Related works on detection defenses and attacks have been comprehensively summarized in recent studies \cite{DBLP:journals/corr/abs-2310-15264,DBLP:journals/corr/abs-2310-14724}. To rigorously evaluate the efficacy of our proposed method, we introduce a paraphrase attack to test the robustness of the detector. Paraphrase attacks were executed using a T5 paraphraser,\footnote{\url{https://huggingface.co/Vamsi/T5_Paraphrase_Paws}} which is designed to reformulate text while retaining its original meaning.

The results in \autoref{table:paraphrased} show that the performance of all methods drops significantly when faced with paraphrased text. Specifically, all zero-shot methods suffer an average performance degradation of 12.93\% AUROC under paraphrase attacks. In contrast, our method demonstrates strong robustness and achieves the most competitive detection performance under paraphrase attacks. Moreover, it maintains robust detection capability across all attack scenarios, with minimal performance degradation.

\section{Samples Distribution Ablation Study}

\begin{figure}[!ht]
    \centering
    \includegraphics[width=0.5\textwidth, trim=0 0 0 0]{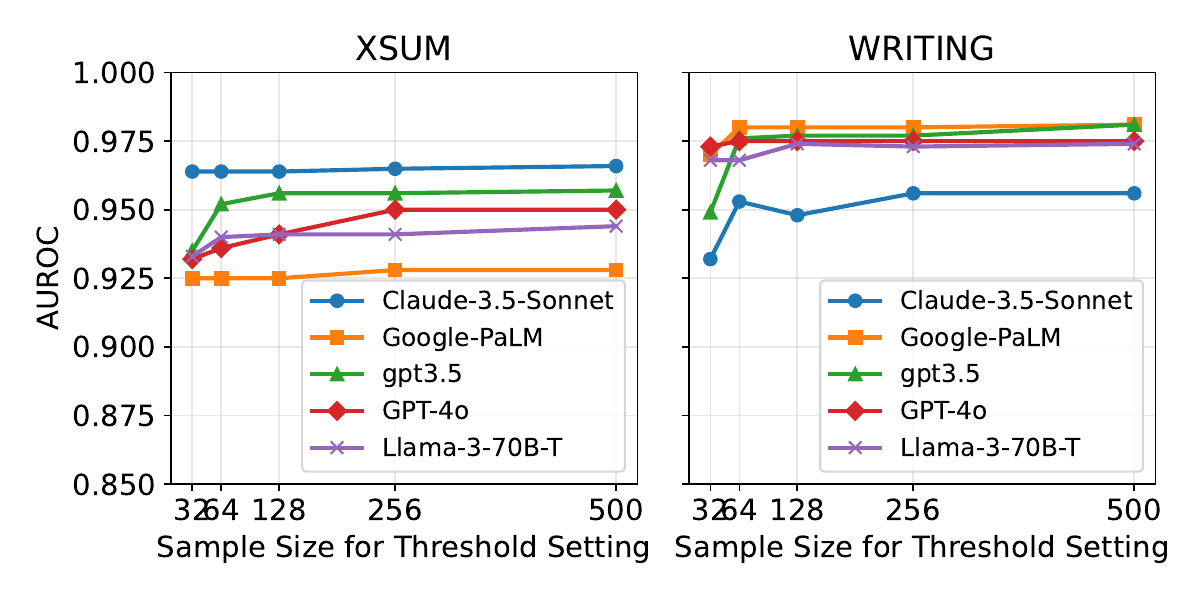}
    \caption{Effect of the Balanced Sample Set Size \textit{n}.}
    \label{fig:balanced_samples}
\end{figure}

We discuss the impact of using different sample sizes \textit{n} to compute the threshold for our method. Specifically, we further validate the robustness of our method under various settings, where the balance between LLM-generated samples and human-written samples is adjusted. Our experimental results on the XSum and Writing Prompts datasets (as shown in \autoref{fig:balanced_samples}) demonstrate that our method achieves stable performance across these scenarios using randomly sampled \textit{n} sample sets. With only 64 samples, our method consistently exceeds 92\% on performance metrics for the XSum dataset and 95\% for the Writing Prompts dataset. Furthermore, when \textit{n} reaches 256, the detector's performance remains nearly constant.

\section{Settings of Similarity Metrics and Text Size Ablation Study}
\label{ablation_settings}

\subsection{Ablation Settings }
We verified the impact of different similarity metrics on our proposed approach. To achieve this, we initially employed an \textit{n}-sentence sliding window technique to augment our dataset. Specifically, we systematically segmented each text sample into all possible complete text fragments using a sliding window mechanism based on \textit{n} sentences. This approach enhanced the diversity of text lengths within our corpus. Subsequently, we conducted a comprehensive analysis of the length distribution of the augmented samples and extracted those that closely matched specific text lengths for further evaluation.

\subsection{Details of Similarity Metrics}
\label{sec:similarity_metrics}
The similarity metrics we used in our study are as follows:

\paragraph{BLEU} 
The Bilingual Evaluation Understudy (BLEU) score~\cite{papineni2002bleu} is a widely used method for assessing the quality of machine translation and is also employed to evaluate the similarity between generated text and reference text in text generation tasks~\cite{haque2022semantic}. The BLEU score calculation is based on the precision of \textit{n}-grams~\cite{manning1999foundations}, which examines the frequency with which n-gram sequences in the generated text appear in the reference text. In our setting, the ``reference'' denotes the text under detection, while the ``hypothesis'' refers to its revised counterpart. The BLEU score ranges from 0 to 1, with 1 indicating a perfect match. A higher BLEU score signifies greater similarity between the original text and its revised version.

\paragraph{TER} 
The Translation Edit Rate (TER) score~\cite{SnoverDSMM06} is a metric primarily used in machine translation evaluation to assess the amount of editing required for a translated text to match a reference translation. Unlike BLEU, which focuses on the precision of matching n-grams, TER calculates the number of edits (insertions, deletions, substitutions, and shifts) needed to transform the hypothesis into one of the references. The TER score ranges from 0 to infinity. A lower TER score is preferable, and a TER score of 0 indicates a perfect match between the hypothesis text and the reference text.

\paragraph{ChrF} 
The Character \textit{n}-gram F-score (chrF) is a metric for machine translation evaluation that calculates the similarity between generated text and reference text using character n-grams instead of word n-grams~\cite{Popovic15}. This method provides a detailed measure of textual similarity at the character level, capturing finer linguistic nuances that word-level metrics may overlook. Similar to the BLEU score, chrF scores range from 0 to 1, with higher chrF scores indicating greater similarity between the generated and reference texts.

\paragraph{Edit Distance} 
Edit Distance~\cite{Navarro01} is a metric used to quantify the dissimilarity between two text strings by calculating the minimum number of operations required to transform one string into the other. These operations typically include insertions, deletions, or substitutions of characters. Unlike metrics that assess similarity or quality of text generation based on semantic or syntactic matches, Edit Distance provides a straightforward, operation-based measure of difference. The score, which can range from 0 upwards, directly corresponds to the number of edit operations needed. A score of 0 indicates identical strings, while higher scores signify greater disparity.

\paragraph{ROUGE} 
The Recall-Oriented Understudy for Gisting Evaluation (ROUGE) score~\cite{lin2004rouge} is a set of metrics designed to evaluate the quality of summaries by comparing them to a set of reference summaries. It is particularly valuable in tasks such as text summarization and text generation, where the goal is to capture the essence of the original text. Similar to the BLEU score, ROUGE scores range from 0 to 1, with 1 indicating a perfect overlap. A higher ROUGE score signifies greater similarity between the generated text and the reference.

\paragraph{GLEU} 
The General Language Understanding Evaluation (GLUE) benchmark~\cite{WangSMHLB18} is a comprehensive suite of resources aimed at the advancement, assessment, and analysis of natural language understanding (NLU) systems. Among these, the STS-B (Semantic Textual Similarity Benchmark) task is designed to evaluate the ability of NLU systems to determine the semantic similarity between pairs of sentences. Systems are required to assign a similarity score ranging from 1 to 5 to each sentence pair, where 1 indicates minimal semantic similarity and 5 denotes a very high degree of semantic similarity.

\paragraph{METEOR} 
The Metric for Evaluation of Translation with Explicit ORdering (METEOR) score~\cite{BanerjeeL05} is an advanced metric designed to address some of the limitations of earlier evaluation methods, such as BLEU, by incorporating a more sophisticated analysis of word-to-word matches between the translated text and the reference translations. It is particularly useful in translation evaluation and other natural language processing tasks requiring an accurate measure of semantic and syntactic alignment. Unlike the BLEU score, METEOR accounts for exact word matches, synonymy, and stemming matches, while introducing a penalty for word order differences. METEOR scores range from 0 to 1, with 1 indicating an exact match between the generated text and the reference.

\paragraph{BLEURT} BLEURT~\cite{DBLP:conf/acl/SellamDP20} is a learning-based evaluation metric based on BERT, specifically designed for assessing English text generation. Traditional evaluation methods, such as BLEU and ROUGE, may not correlate well with human judgment. BLEURT addresses this by modeling human judgment using a few thousand potentially biased training examples. A higher BLEURT score signifies greater similarity between the original text and its revised version.

\paragraph{BERTScore} BERTScore~\cite{DBLP:conf/iclr/ZhangKWWA20} is a metric for evaluating the quality of text generation. It compares the similarity between candidate and reference texts using contextual embeddings from BERT, a pre-trained language model. Unlike traditional metrics (e.g., BLEU), which rely on exact word matches, BERTScore captures semantic meaning by calculating cosine similarity between embeddings. This enables it to account for synonyms and variations in phrasing, making it a robust tool for assessing natural language generation tasks.

\paragraph{BARTScore} 
BARTScore~\cite{yuan2021bartscore} is an evaluation metric designed for natural language generation tasks like summarization and translation. It leverages the BART model to assess the quality of generated text by comparing it to reference texts and measuring their similarity. This approach is versatile, allowing for multi-dimensional evaluations by considering different directions of similarity, such as from generated to reference text. It functions in an unsupervised manner, meaning it does not require additional annotated data, making it an efficient and adaptable tool for assessing the quality and relevance of generated content in various applications.

\end{document}